\documentclass[journal,a4paper,11pt]{IEEEtran}

\pdfoutput=1
\usepackage{epsfig}
\usepackage{graphics}
\usepackage{cite}
\usepackage{amsmath}

\usepackage{caption}
\usepackage{floatrow}

\usepackage{verbatim}

\usepackage{array}

\begin{document}

\title{Opinion mining of text documents written in Macedonian language}%

\author{Andrej Gajduk$^*$~and~Ljupco Kocarev$^{* \ddagger \S}$
\thanks{$^*$ Macedonian Academy of Sciences and Arts.}
\thanks{$^\ddagger$ Ss. Cyril and Methodius University, Faculty of Computer Science and Engineering, Skopje.}
\thanks{$^\S$ BioCircuits Institute, University of California at San Diego.}}

\maketitle


\begin{abstract}
The ability to extract public opinion from web portals such as review sites, social networks and blogs will enable companies and individuals to form a view, an attitude and make decisions without having to do lengthy and costly researches and surveys. In this paper machine learning techniques are used for determining the polarity of forum posts on kajgana which are written in Macedonian language. The posts are classified as being positive, negative or neutral. We test different feature metrics and classifiers and provide detailed evaluation of their participation in improving the overall performance on a manually generated dataset. By achieving 92\% accuracy, we show that the performance of systems for automated opinion mining is comparable to a human evaluator, thus making it a viable option for text data analysis. Finally, we present a few statistics derived from the forum posts using the developed system.
\end{abstract}

\section{Introduction}
The World Wide Web (Web) has tremendously influenced our lives by changing the way we manage and share the information. Today, we are not only static observers and receivers of information, but in turn, we actively change the information content and/or generate new pieces of information. In this way, the entire community becomes a writer, in addition to being a reader. Different mediums, such as blogs, wikis, forums and social networks, exist in which we can express ourselves by posting information and giving opinion on various subjects, ranging from politics and health to product reviews and travelling. 

	Sentiment analysis (also referred as opinion mining) concerns application of natural language processing, computational linguistics, and text analytics to identify and extract subjective information in source materials. Opinion Mining operates at the level of documents, that is, pieces of text of varying sizes and formats, e.g., web pages, blog posts, comments, or product reviews. We assume that each document discusses at least one topic, that is, a named entity, event, or abstract concept that is mentioned in a document. Sentiment is the author’s attitude, opinion, or emotion expressed on a topic. Although sentiments are expressed in natural language, they can in some cases be translated to a numerical or other scale, which facilitates further processing and analysis. Since the palette of human emotions is so vast and it is hard to select even the basic ones, most of the authors in the NLP community work with representation of sentiments according to their polarity, which means positive or negative evaluation of the meaning of the sentiment. 
	
	It is now well-documented that the opinions/views expressed on the web can be influential to readers in forming their opinions on some topic \cite{lin2012sentiment}, and therefore, they are an important factor taken into consideration by product vendors \cite{pang2002thumbs} and policy makers \cite{horrigan2008online}. There exists evidence that this process has significant economic effects \cite{antweiler2004all,archak2007show,chevalier2003effect}. Moreover, the opinions aggregated at a large scale may reflect political preferences \cite{mullen2006preliminary,tumasjan2010predicting} and even improve stock market prediction  \cite{bollen2011twitter}. For the recent surveys on sentiment analysis or opinion mining we refer readers to \cite{pang2008opinion,tang2009survey,tsytsarau2012survey}.
	
	The outline of the paper is as follows. In Section 2 the problem of opinion mining is formally defined. The proposed approach is outlined in Section 3. In Section 4 we give details about the datasets used in our experiments. In Section 5 the performance achieved using the different feature representation, classifiers and other text processing techniques is compared. A few statistics on the forum posts on kajgana derived using opinion mining are presented in Section 6. Section 7 concludes this paper.

\section{Problem Definition}

In our experiment, we accept the classification of opinions according to their polarity i.e. polarity classification, used by the majority of authors \cite{pang2002thumbs,pang2008opinion}. Pang and Lee \cite{pang2008opinion} define polarity as the point on the evaluation scale that corresponds to our \textit{positive} or \textit{negative} evaluation of the meaning of the expressed opinion. However, not all texts are opinionated, so the method proposed by \cite{godbole2007large} which rates subjectivity and polarity separately is used.
The problem is defined as follows:  \\
	 \emph{Given a piece of text, decide whether it is subjective or objective, then assuming that the overall opinion in it is about one single issue or item, classify the opinion in subjective posts as falling under one of the two categories: positive or negative.} 

\section{Proposed approach}

\subsection{Data representation}

Text data in machine learning is commonly represented by using the bag-of-features method~\cite{nakagawa2010dependency,read2005using,paul2010summarizing,rushdi2011experiments}. This method is described as follows: let $D = \lbrace f_1, \ldots ,f_m \rbrace$ be a predefined set of $m$ features that can appear in a forum post. We will refer to $D$ as a feature dictionary. The features in the dictionary can be unigrams i.e. words such as \textit{“great”} and \textit{“wasteful”}, bigrams i.e. word pairs such as \textit{“not comfortable”} or n-grams in the general case. Every post is represented by a vector of real numbers which correspond to a single feature in the feature dictionary. These values are computed using four different feature metrics. \

\begin{itemize}
\item n-gram presence \\
\begin{equation}
\label{presence}
presence_i^p = \left\{
	\begin{array}{ll}
		1,  & \mbox{if } t_i^p \neq 0 \\
		0, & \mbox{otherwise}
	\end{array}
\right.
\end{equation}
\item n-gram count \\
\begin{equation}
\label{count}
count_i^p = t_i^p
\end{equation}
\item n-gram frequency \\
\begin{equation}
\label{freq}
freq_i^p = \frac{t_i^p}{\sum_{j} t_j^p}
\end{equation}

\item n-gram frequency-inverse document frequency \\
\begin{equation}
\label{itdf}
ifreq_i^p = freq_i^p \log \frac{||P||}{||P_i||}
\end{equation}
\end{itemize}

In (\ref{presence}--\ref{itdf}) $t_i^p$ is the number of occurrences of the $i$th n-gram in the post $p$, $P$ is a set of all the posts and $P_i$ is a set of posts containing at least one occurrence of the $i$th n-gram. Unigrams are the most commonly used in text mining, although some authors \cite{zeng2013approach} recommend using bigrams. Their arguments include dealing with word negation and emphasizing which are very important in the domain of polarity classification.

\subsection{Classifiers}

The proposed feature metrics are evaluated using the two classifiers preferred by the majority of researchers in text classification~\cite{mullen2004sentiment,ye2005sentiment,gamon2004sentiment,koppel2006importance}.  \\
\begin{itemize}
\item Support Vector Machines \\
\item Naive Bayes \\
\end{itemize}
As discussed earlier, the classification will take place in two phases. First, subjectivity classification is performed where the comment is rated as either subjective or objective. Then if the post is subjective, it is classified as being either positive or negative. The latter will be referred to as polarity classification.

\subsection{Preprocessing}

Stop Words: Filtering stop words is a common practice in text mining~\cite{wiebe2005annotating,chaovalit2005movie,strapparava2007semeval,esuli2006determining}. Stop words are words with no informational value, such as function and lexical words. A suitable list of stop words in Macedonian language is difficult to obtain so one had to be manually prepared for this experiment. The list of stop words constitutes of 170 entries. 

Stemming: Stemming has been extensively used to increase the performance of information retrieval systems for many international languages such as: English, French, Portuguese, to name a few \cite{frakes2003strength,savoy2008searching}. Stemming is a technique which aims to reduce a word to its stem or root form. Thus, literally different words that share a common stem may be abstracted as a single informational entity. There are several common approaches to stemming as categorized in \cite{sharma2012stemming} namely affix removal method, successor variety method, n-gram method and table lookup method. Affix removal which includes algorithms such as Lovins or Porter, is the most popular method, but relies heavily on manually defined rule sets. A good rule set for Macedonian is yet to be defined, which is why we decided to use a stemming method that relies on nothing more but the set of words that need to be stemmed.  This method is called peak-and-plateau and is based on tries. For a more detailed explanation to this method we refer readers to \cite{hafer1974word}.

\subsection{Rule bigrams}
Some authors propose a different way of incorporating bigrams into the feature vector \cite{ghiassi2013twitter,socher2013recursive,wang2012baselines,kang2012senti} which will be refereed to as \textit{rule bigrams}. According to this approach all negatory words are appended a tag \textit{e.g.} \textit{“not”} to the word following the negatory word in the sentence. Thus the bigram \textit{“not good”} becomes the unigram \textit{“notgood”}. This method is adopted and expanded to emphasizery words, thus transforming bigrams such as \textit{“most disgusting”} and \textit{“very disgusting”} into the same unigram \textit{e.g.} \textit{“verydisgusting”}. This approach is adequate when using unigram presence as a feature vector, but we propose an alteration when applying it in combination with other feature metrics that rely on counting the unigram occurrences. Any occurrence of an unigram preceded by an emphasizing word is counted as two occurrences of the corresponding unigram i.e. $\hat{t}_i^p  ̂=2 t_i^p$, whereas any occurrence of an unigram preceded by a negatory word is considered as -1 occurrence of the corresponding unigram i.e. $\hat{t}_i^p =-t_i^p$.

\section{Dataset}

The domain used in this study is forum posts which are written in Macedonian language from the kajgana forum. Forum posts tend to be less focused and organized than other text documents such as product reviews for instance, and consist predominantly of informal text. The posts on kajgana are grouped into 47 disjoint topics which are then divided into subtopics (over 50,000) and are 60 words long on average. There are a total of 4 million unique words in the posts. In our experiment, we ignored words that have less than 5 occurrences in order to reduce the total dictionary size and to eliminate type errors. This left us with 800,000 unique words.
A total of 800 posts were manually tagged of which 260 are positive, 260 are negative and 280 are objective posts. This dataset will be used for evaluations on the different classifiers and feature representations. All evaluations are done using 10-fold cross validation to avoid over-fitting.

\section{Results}

First, the aforementioned feature representations using unigrams in combination with the two proposed classifiers are evaluated. Inverse frequency the best feature representation  followed by presence~(Table.~\ref{tab:un_no_preproc}). As for classifiers, SVM outperforms NB on every feature representation.

\begin{table}
\caption{Accuracy, no preprocessing} \label{tab:un_no_preproc}
\begin{tabular}{ l | r | r }
  Accuracy & SVM & NB \\
  \hline
    Presence	& 0.76	& 0.64 \\
	Count	    & 0.73  & 0.55 \\
	Frequency	& 0.72	& 0.61 \\ 
	IFrequency  & 0.94	& 0.78 \\
\end{tabular}
\end{table}

Surprisingly, stemming and stop words removal reduces accuracy~(Table~\ref{tab:stop_words}). More specifically the accuracy drops from 0.94 to 0.74 when using an SVM classifier and from 0.78 to 0.62 when using an NB classifier. One possible reason is that the word stemming algorithms does not perform well for the Macedonian language.

\begin{table}
\caption{Accuracy, with preprocessing} \label{tab:stop_words}
\begin{tabular}{ l | r | r }
  Accuracy & SVM & NB \\
  \hline
    Presence	& 0.76	& 0.63 \\
	Count	    & 0.72  & 0.56 \\
	Frequency	& 0.70	& 0.60 \\ 
	IFrequency  & 0.74	& 0.62 \\
\end{tabular}
\end{table}

As mentioned earlier the proposed feature representations can be applied to n-grams of any size, although so far only unigrams have been used. Next, we evaluate presence and ifrequency using bigrams, alone and in combination with unigrams~(Table~\ref{tab:simple_bigrams}). Bigrams alone are not good features, but when used in conjunction with unigrams they show a slight improvement when presence as feature representation is used from 0.76 to 0.78 with SVM and from 0.63 to 0.67 with NB.

\begin{table}
\caption{Accuracy, bigrams} \label{tab:simple_bigrams}
\begin{tabular}{ l | r | r }
  Presence & SVM & NB \\
  \hline
  Unigrams only &	0.76 &	0.63 \\
   Bigrams only	& 0.54 &	0.52 \\
Unigrams bigrams	& 0.79	 & 0.67 
\vspace{4ex}
\\
IFrequency	& SVM &	NB \\
\hline
Unigrams only &	0.74 &	0.62 \\ 
Bigrams only &	0.55 &	0.52 \\
Unigrams bigrams & 	0.75	 & 0.62 \\
\end{tabular}
\end{table}

Finally, in Table~\ref{tab:rule_bigrams} the accuracy when using rule bigrams (only negation rules, only emphasis rules and both together) are given. The results show that rule bigrams do not impact classification accuracy, with the exception of negation rules that achieves a slight increase in accuracy for unigram presence .

\begin{figure*}
\label{fig:mood_topic}
\includegraphics[scale=0.7]{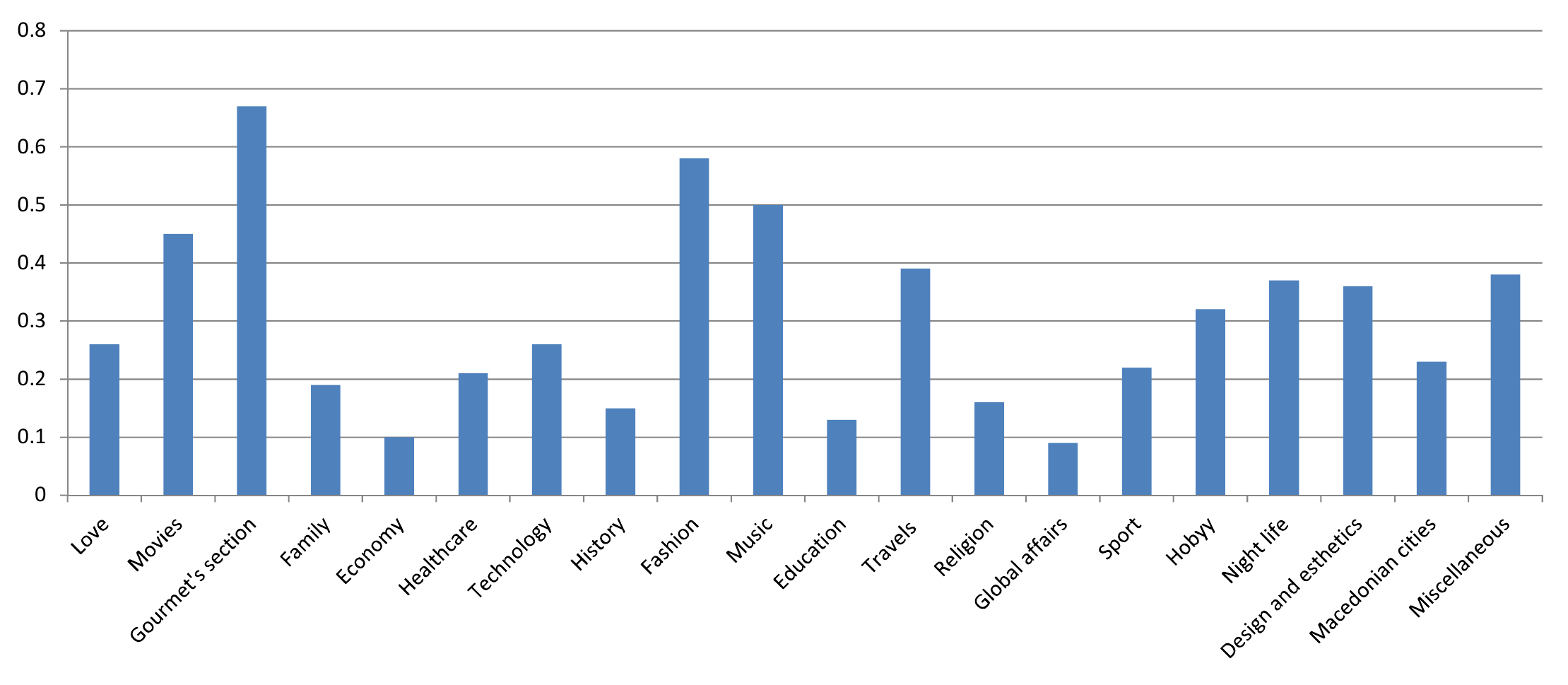}
\caption{Mood by topic}
\end{figure*}

\begin{table}
\caption{Accuracy, rule  bigrams} \label{tab:rule_bigrams}
\begin{tabular}{ l | r | r }
  Presence & SVM & NB \\
  \hline
  Unigram &	0.76	 & 0.63 \\
Negations only &	0.78 &	0.62 \\
Emphasizers only &	0.76 &	0.61 \\
Both	& 0.77	& 0.62 
\vspace{4ex}
\\
IFrequency	& SVM &	NB \\
\hline
Unigram &	0.74 &	0.62 \\
Negations only	 & 0.73 &	0.59 \\
Emphasizers only &	0.74 &	0.59 \\
Both &	0.74 &	0.61 \\
\end{tabular}
\end{table}

\section{Statistics}

\captionsetup[figure]{justification=raggedright}

Using the combination of unigram ifrequency for a feature representation and SVM as a classifier some interesting properties of forum posts in general can be demonstrated. As stated above the forum posts are divided into several topics. Let us denote with $p_t$ the number of positive posts and with $n_t$ the number of negative posts for each topic $t$. The overall mood on the topic $m_t$ is defined as 
\begin{equation}
\label{eq:mood}
m_t = \frac{p_t}{p_t+n_t}
\end{equation}
Interestingly, people are most positive when discussing food (Gourmet’s section) and fashion, but are extremely negative on global affairs and the economy~(Fig.~1). .

In a similar fashion the  posts can be grouped and their mood calculated by month as displayed in Fig.~2. The public mood is highest in spring (May and April), probably due to the good weather during these two months.

\begin{figure*}
\label{fig:mood_month}
\includegraphics[scale=0.5]{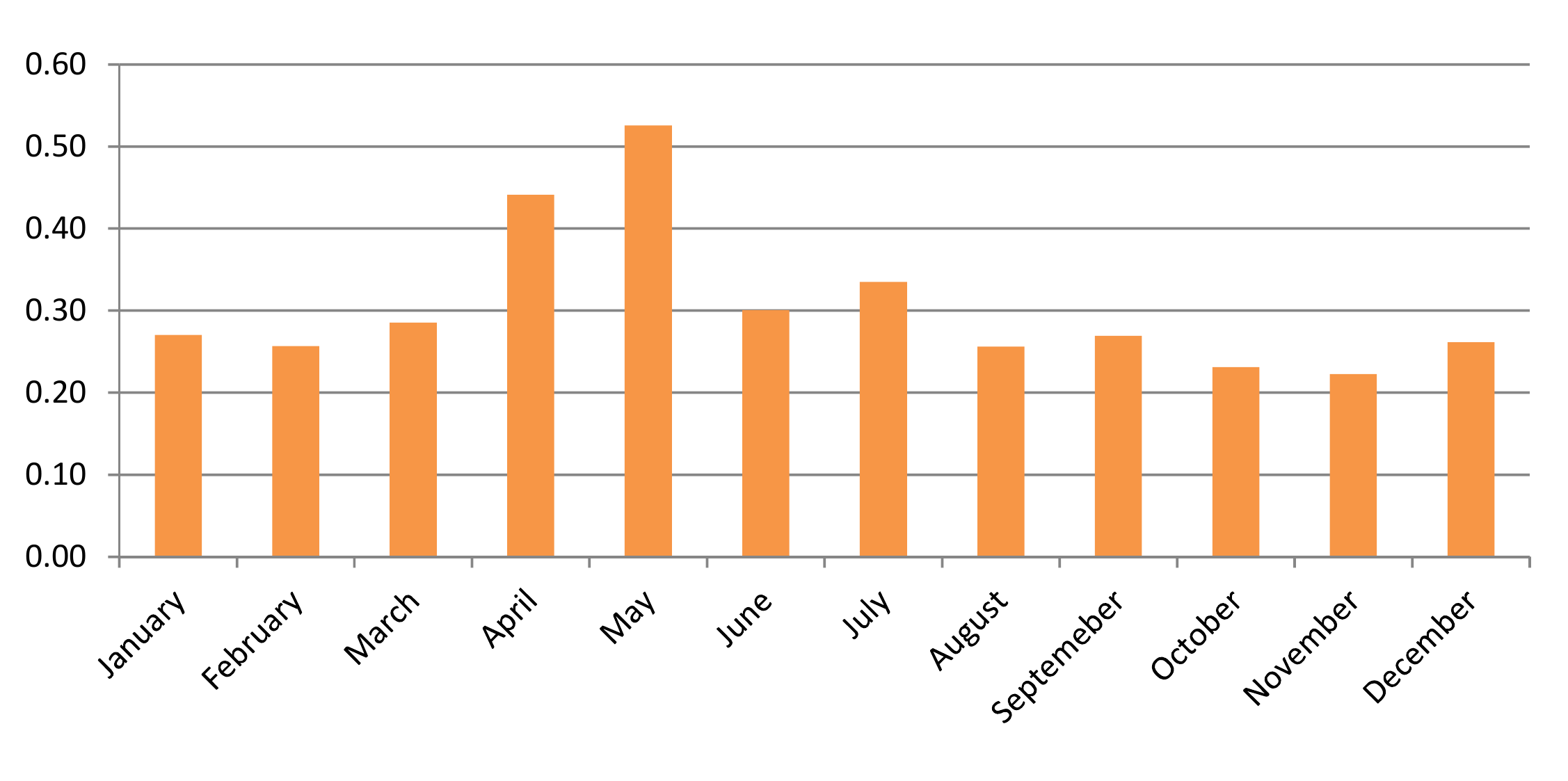}
\caption{Mood by month}
\end{figure*}

\section{Conclusion}

In this paper forum posts written in Macedonian language are labeled as being positive, negative or objective. We show that this can be done with great accuracy using simple text feature extraction metrics such as unigram presence and standard classifiers such as Naive Bayes. The best accuracy is achieved by using a combination of unigram frequency-inverse document frequency for a feature metrics and support vector machines as a classifier: 0.96 on subjectivity classification, 0.96 on polarity classification or a total classification accuracy of 0.92. Additionally, we tested various techniques for improving the performance. Of these, word stemming and stop words removal had a negative effect on classification accuracy. The use of bigrams does not help with the classification task while using rule bigrams increases the accuracy only slightly in polarity classification.

\end{document}